\documentclass[10pt, a4paper]{article}
\usepackage{lrec}
\usepackage{multibib}
\newcites{languageresource}{Language Resources}
\usepackage{graphicx}
\usepackage{tabularx}
\usepackage{soul}
\usepackage{amsmath}
\usepackage{cuted}

\usepackage{epstopdf}
\usepackage[utf8]{inputenc}

\usepackage{hyperref}
\usepackage{xstring}

\usepackage{color}

\graphicspath{{graphics/}}
\newcommand\blfootnote[1]{%
  \begingroup
  \renewcommand\thefootnote{}\footnote{#1}%
  \addtocounter{footnote}{-1}%
  \endgroup
}

\title{Seeing the Forest and the Trees: Detection and Cross-Document Coreference Resolution of Militarized Interstate Disputes}

\name{Benjamin J. Radford}

\address{Department of Political Science and Public Administration \\
		University of North Carolina at Charlotte \\
         benjamin.radford@uncc.edu\\}

\abstract{
Previous efforts to automate the detection of social and political events in text have primarily focused on identifying events described within single sentences or documents. Within a corpus of documents, these automated systems are unable to link event references---recognize singular events across multiple sentences or documents. A separate literature in computational linguistics on event coreference resolution attempts to link known events to one another within (and across) documents. I provide a data set for evaluating methods to identify certain political events in text and to link related texts to one another based on shared events. The data set, \emph{Headlines of War}, is built on the \emph{Militarized Interstate Disputes} data set and offers headlines classified by dispute status and headline pairs labeled with coreference indicators. Additionally, I introduce a model capable of accomplishing both tasks. The multi-task convolutional neural network is shown to be capable of recognizing events and event coreferences given the headlines' texts and publication dates.\\ 
\newline  
}

\begin{document}

\maketitleabstract

\section{Introduction}

The automation of political event detection in text has been of interest to political scientists for over two decades. \newcite{schrodt:1998} introduced KEDS, the Kansas Event Data System in the 1990s, an early piece of event coding software. Successors to KEDS include TABARI, JABARI-NLP, and now PETRARCH in its various incarnations \cite{schrodt:2009,schrodt:etal:2014}. However, these tools rely primarily on performing pattern-matching within texts against dictionaries, limiting their ability to recognize singular events across multiple sentences or documents. This leads to unwanted duplication within event data sets and limits the types of detected events to those that are concisely summarized in a single line.\blfootnote{The data set described in this paper is available on Harvard Dataverse: \url{https://doi.org/10.7910/DVN/8TEG5R}.}

Social scientists have recently begun exploring machine learning-based approaches to coding particular types of political events \cite{beieler:2016,protestnews:2019,radford:2019}. However, these efforts still mainly focus on classifying events at the sentence or document level. In this paper, I propose an approach to event-coding that is able to detect singular events at both the document (headline) level as well as across documents. Therefore, this challenge is not only a classification task but also a coreference prediction task; headlines are classified as pertaining to events and multiple headlines referring to the same event are identified as coreferencing the event.

This two-part challenge mirrors real-world cross-document event coreference detection. The first task is the identification of relevant events among a corpus that contains relevant (positive) and irrelevant (negative) events. The second task is to identify event coreferences across documents. Multiple articles may refer to the same event, and there may be an arbitrary number of distinct events within the corpus. This second task is conceptualized as link prediction wherein a link between articles signifies that they refer to the same event. 

Event linking, or coreference resolution, has been studied in the context of computer science and computational linguistics. This research is often framed within the larger problem of automated knowledge base population from text. \newcite{lu:ng:2018} provide a review of research in this area over the previous two decades including discussion of standard data sets, evaluation methods, common linguistic features used for coreference resolution, and coreference resolution models. Notable datasets for coreference resolution include one built by \newcite{hong:etal:2016} using the Automated Content Extraction (ACE2005\footnote{\url{http://projects.ldc.upenn.edu/ace}}) corpus, a data set produced by \newcite{song:etal:2018} in support of the Text Analysis Conference Knowledge Base Population effort, and the EventCorefBank (ECB) and ECB+\footnote{\url{http://www.newsreader-project.eu/results/data/the-ecb-corpus/}} data sets \cite{bejan:harabagiu:2010,cybulska:vossen:2014}.

Advances in event linking also promise to enhance automated event data generation for social science applications. Event data sets like ICEWS, GDELT, and Pheonix suffer from duplicate event records when single events are reported multiple times by multiple sources \cite{boschee:etal:2015,leetaru:schrodt:2013,althaus:etal:2019}. Typically, duplicated records are removed via heuristics based on the uniqueness of event attribute sets. Event linking techniques may allow event data sets like to these to better represent complex phenomena (e.g., wars) that are described across multiple documents while avoiding the duplication problem.

The paper proceeds as follows. I first describe a novel data set designed to evaluate performance on cross-document event detection. I then introduce a model capable of both event detection and cross-document coreference prediction and evaluate its performance on out-of-sample data. The paper concludes with a discussion of the limitations of the evaluation data set and suggested directions for future research.

\section{Data}
I introduce here a task-specific evaluation data set referred to as the \emph{Headlines of War} (HoW) data set. HoW takes the form of a node list that describes news story headlines and an edge list that represents coreference links between headlines. HoW draws headline and coreference data from two sources. The first is the \emph{Militarized Interstate Disputes} data set (MIDS) version 3. MIDS provides a set of newspaper headlines that coreference interstate disputes. \emph{The New York Times} (NYT) provides a second source of headlines that constitute the negative (non-coreferential) samples.

\subsection{MIDS}
MIDS is a standard in political science and international relations.\footnote{The \emph{Militarized Interstate Disputes} data set will be referred to as MIDS while an individual dispute will be referred to as MID (plural: MIDs). A MID incident will sometimes be referred to as MIDI.} It is published by the Correlates of War Project, an effort that dates to 1963 \cite{singer:small:1966}. A MID is a collection of ``incidents involving the deliberate, overt, government-sanctioned, and government-directed threat, display, or use of force between two or more states'' \cite{maoz:etal:2019}. As such, many MIDs, and the incidents they comprise, are macro-level events that may occur over an extended period of time and comprise many smaller events. For example, a number of ceasefire violations in Croatia in February, 1992, together constitute incident 3555003. 3555003 is one of many incidents that make up MID 3555, the Croatian War for Independence. MIDs and the incidents they comprise tend to be larger-scale than the events found in typical event data sets. 

MIDS differs from automated event data in several ways. Automated event data sets (referred to herein simply as ``event data'') like GDELT, ICEWS, and Phoenix typically document discrete events that are easily described in a single sentence. This is due, in part, to the fact that the necessary coding software parses stories sentence-by-sentence and uses pattern-matching to identify the key components of an event within a given sentence. This leads to data sets that feature simple events and often include duplicate records of events. Failure to deduplicate led, in one case, to an incident in which a popular blog was forced to issue corrections due to the over-counting of kidnapping events in GDELT \cite{chalabi:2014}.

Because it is coded manually, MIDS features more complex events than automated event data systems are capable of producing. MIDs comprise incidents, and incidents may (or may not) themselves comprise a number of actions that would each constitute their own entry in an automated event data set. Because each MID is coded from a number of news sources, duplication of disputes is not a concern; human coders are capable of mapping stories from multiple news sources to the single incident or dispute to which they all refer.

MIDS provides HoW with positive class labels (i.e.,, headlines associated with MIDs) and positive coreferences (pairs of headlines associated with common MIDs). I use the third version of MIDS due to the availability of a subset of the source headlines used to produce the data set \cite{ghosn:etal:2004}.\footnote{The source data are available at \url{https://correlatesofwar.org/data-sets/MIDs}. An effort to update HoW with MIDS version 4 headlines is underway.}

\begin{figure}
\begin{center}
\includegraphics[width=\linewidth]{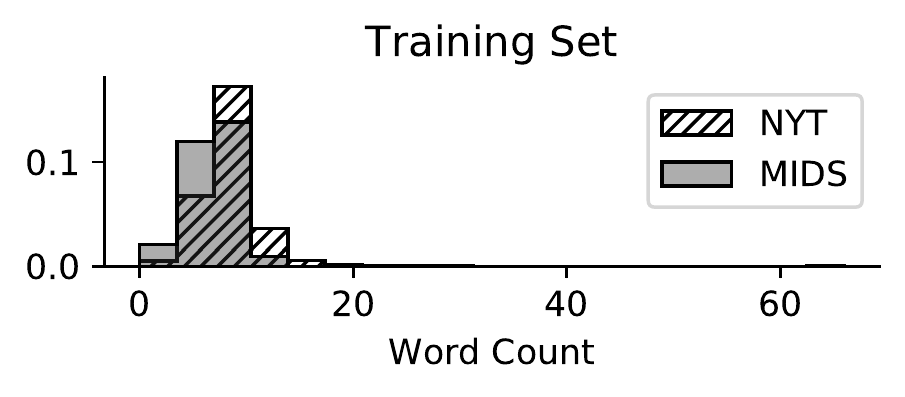} \\
\includegraphics[width=\linewidth]{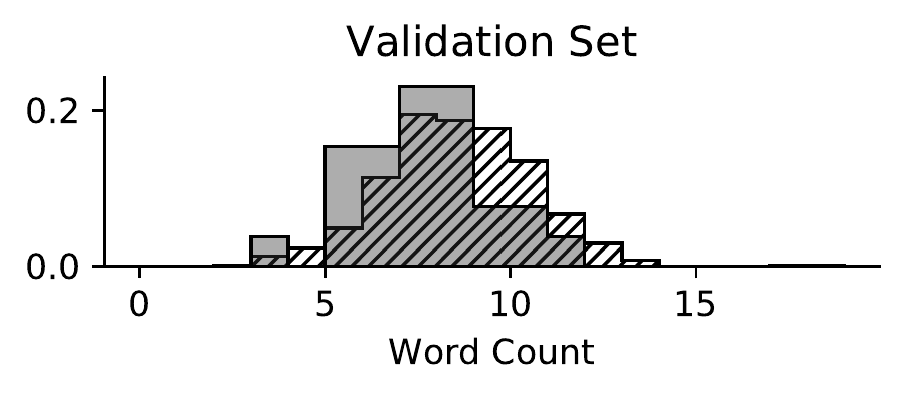} \\
\includegraphics[width=\linewidth]{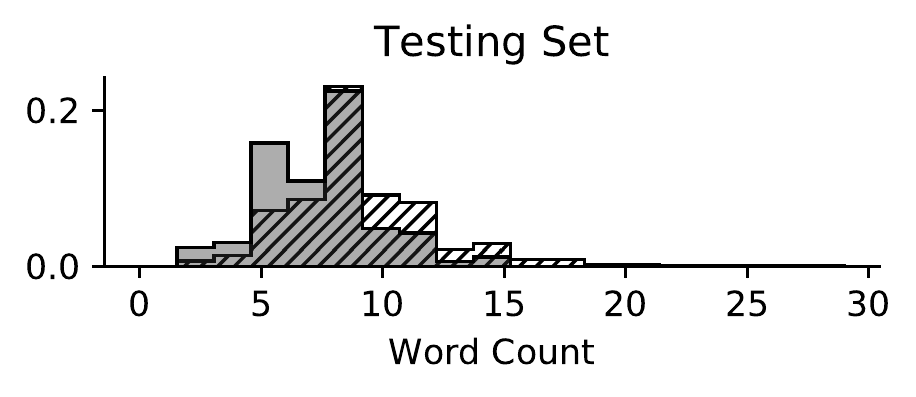}
\caption{Sentence length in words by HoW subset.}\label{fig:sentence_length}
\end{center}
\end{figure}

\subsection{The New York Times}
Negative samples, headlines not associated with militarized interstate disputes, are drawn from \emph{The New York Times} for the same period as that covered by MIDs 3.0: 1992--2001.\footnote{MIDs 3.0 only includes those conflicts from 1992 that were ongoing in 1993. For simplicity, NYT headlines are sampled from January 1, 1992.} NYT headlines and their associated sections (e.g., World, US, Sports, ...) are available from https://spiderbites.nytimes.com. HoW contains only samples from the World section. This is to ensure that the resulting task is sufficiently difficult. Articles drawn from the World section are more likely to mirror the MIDs headlines in tone and substance; distinguishing between MIDS headlines and NYT World headlines should, therefore, be more difficult than it would be if articles from all sections were sampled.

\begin{table}
\begin{center}
\begin{tabularx}{\linewidth}{Xrrr}
\hline \hline 
& Training & Validation & Testing \\
\hline
 Start date (01/01) & 1992 & 1997 & 1998 \\ 
 End date (12/31) & 1996 & 1997 & 2001 \\
Headlines & 4,987 & 966 & 13,515 \\ 
 MID headlines & 123 & 26 & 108 \\ 
 $\neg$MID headlines & 4,864 & 940 & 13,407 \\ 
 Characters & 249,092 & 47,018 & 756,230 \\ 
 Unique MIDs & 10 & 6 & 3 \\ 
 Unique Incidents & 26 & 6 & 4 \\ 
 Links & 3,378 & 678 & 30,342 \\ 
 Positive links & 563 & 113 & 5,057 \\ 
 Negative links & 2,815 & 565 & 25,285 \\ 
 \hline \hline
\end{tabularx}
\caption{Summary statistics of HoW data set partitions.}\label{tab:datastats}
\end{center}
\end{table}

\subsection{Putting it Together: HoW}
The HoW data are partitioned into three parts: training, validation, and testing. Partitioning is performed by year to make it unlikely that a single MID incident's reference headlines are found across all three partitions. An unfortunate consequence of doing so is that it is difficult to control the relative sizes of each partition. MID incidents are not evenly distributed across years, and so the validation set is smaller (in terms of headline-pairs) than the training set, which is, in turn, smaller than the testing set.

Summary statistics for each partition of HoW are given in Table~\ref{tab:datastats}. Not all MIDs and MID incidents during the relevant time periods are included. This is due to the fact that the MIDS source data do not report headlines for all incidents. In many cases, page numbers and sections numbers are provided in lieu of the headline text itself. Therefore, HoW contains a total of only 18 unique MIDs (with one appearing in two partitions) and 36 unique incidents.

Each partition comprises a node list and an edge list. The node list contains the headline text, publication date, associated MID identifier and incident identifier (if applicable), and an indicator of whether the headline is a positive (MID) sample or a negative (NYT World) sample. The edge list includes positive links between headlines if they refer to the same MID incident along with a sample of negative links drawn randomly from NYT World and MIDS headlines. Therefore, a single MID incident is represented in the edge list by a fully-connected subgraph of headlines.

Figure~\ref{fig:sentence_length} depicts the distribution of headline lengths, in words, for each of the HoW subsets. The average headline length is just under nine words.

\begin{figure}
\begin{center}
\includegraphics[width=\linewidth]{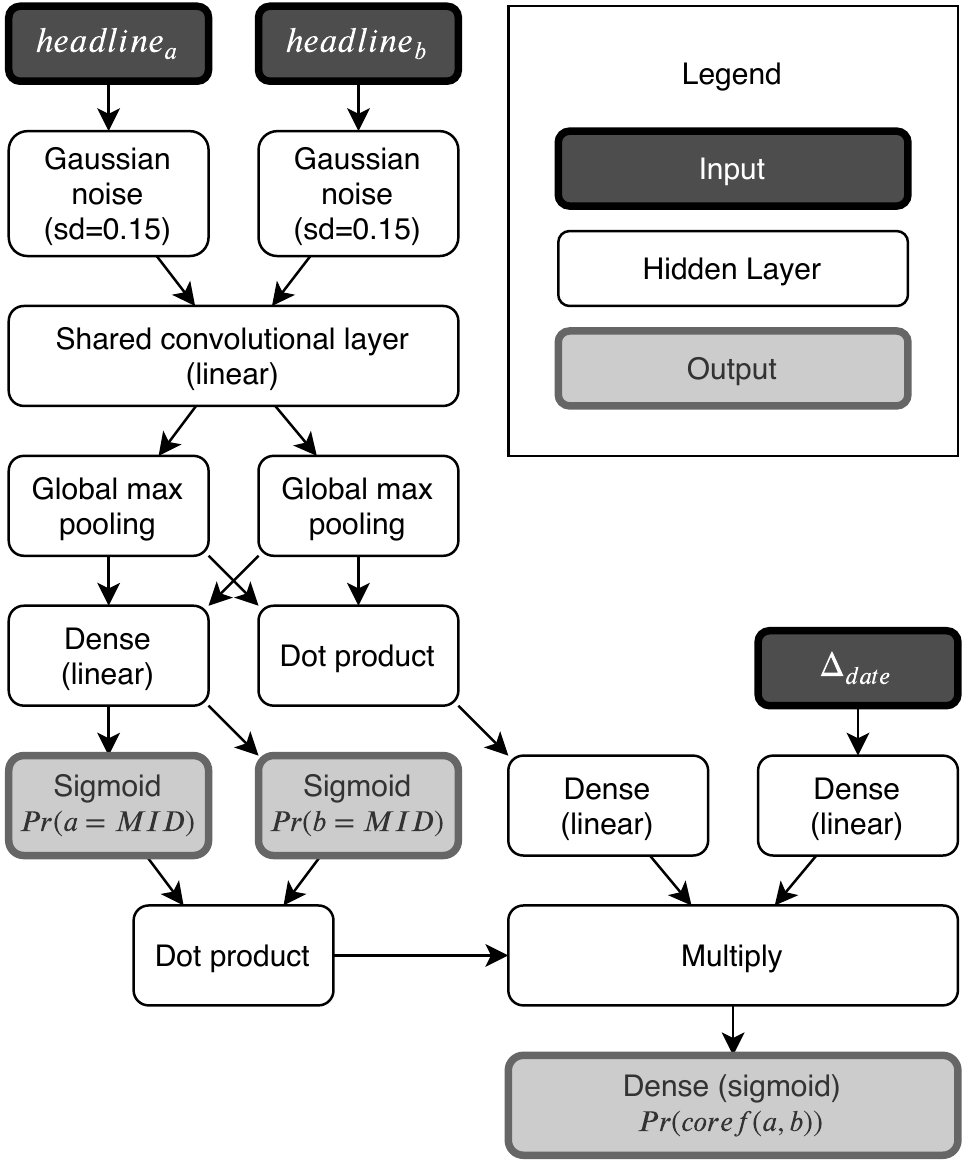}
\end{center}
\caption{Model architecture for headline classification and corefence prediction.}\label{fig:architecture}
\end{figure}

\section{Modeling Strategy}
To demonstrate that HoW presents a tractable pair of tasks, I describe a model capable of accomplishing, to a degree, both headline classification and link prediction on the data set. The model is a multi-task neural network that takes as input numerical representations of two headlines and the reciprocal of $1+(\Delta publication dates)$. The model then predicts the MID status of both headlines, $headline_a$ and $headline_b$, and whether or not the headlines refer to the same MID incident. 

\subsection{Preparing the Headlines}
The first step of modeling is to remove all punctuation from the headlines' texts. For convenience, headlines are zero-padded such that they are all of equal length. Headlines are then tokenized and word vectors are substituted for each word.\footnote{When a word vector cannot be obtained for a given token, that token is simply dropped.} Pre-trained word vectors are obtained from Facebook's fastText \cite{mikolov:2018}. FastText is selected because it is able to produce word vectors for out-of-sample words---those that it has not previously seen. Word vectors are length 300 real-valued vectors that represent words in such a way that semantically and syntactically related words share similar vectors.

\subsection{Model Architecture}
The model itself comprises a single convolutional layer of size $300 \times 15 \times 3$ and three dense, fully-connected layers for predicting MID status and coreference status. For a given input pair, the model outputs three predictions: 
\begin{align*}
& Pr\left(a=MID | headline_a \right) \\
& Pr\left(b=MID | headline_b \right) \\
& Pr\left( coref(a,b) | headline_a, headline_b, \Delta_{date} \right)
\end{align*}
where $coref(a,b)$ indicates that $headline_a$ and $headline_b$ refer to the same MID incident and $\Delta_{date} = 1/(1+|date_a - date_b|)$. The overall model architecture is depicted in Figure~\ref{fig:architecture}. The model contains 13,537 trainable parameters, 13,515 of which are in the convolutional layer.

The intuition behind the model is as follows. MID classification should be the same task regardless of whether the input headline is $a$ or $b$. Therefore, the convolutional layer and subsequent densely-connected layer are shared between the two. Combined, this outputs a predicted probability that a given headline describes a MID incident. After the convolutional layer and an element-wise maximum value pooling layer, the dot product of the hidden states representing $headline_a$ and $headline_b$ is computed; this represents the similarity of the two headlines. This value is multiplied by the predicted probabilities that each headline represents a MID incident as well as by a linear function of the time difference (in days) between the two headlines. A sigmoid activation is applied to this product; this value represents the probability of a MID incident coreference between $headline_a$ and $headline_b$. Therefore, MID incident coreferences are most likely when the model predicts that both $headline_a$ and $headline_b$ describe MID incidents, when the hidden state representations of those headlines are most similar, and when the publication date difference between the headlines is small.

\subsection{Training Procedure}
The model is trained for 100 epochs on batches of 64 training samples. The validation set is used for parameter tuning. The testing set remains unobserved until the final model is selected. Because the model must predict three binary responses, the loss function is the unweighted sum of the three binary cross-entropy terms given in Equation~\ref{eq:loss}. The model is fit using Nadam, a variant of the Adam optimizer with Nesterov momentum \cite{dozat:2016}.
\begin{align}
\label{eq:loss}
\begin{split}
    Loss = & - \textstyle \sum_{i=0}^{1} y_{i}^{headline_a} \text{log}\left( Pr(a=i) \right) \\
     & -  \textstyle \sum_{j=0}^{1}  y_{j}^{headline_b} \text{log}\left( Pr(b=j) \right) \\
     & - \textstyle \sum_{k=0}^{1} y_{k}^{coref(a,b)} \text{log}\left( Pr(coref(a,b)=k) \right) \\
\end{split}
\end{align}

This model is similar in some aspects to the one introduced by \newcite{krause:etal:2016}. Major differences include the use of fastText vectors here rather than word2vec vectors, the requirement in this model that it not only identifies coreferential headlines but also that it discriminates between events and non-events, and the lack of additional contextual information about event pairs.\footnote{\newcite{krause:etal:2016} include type compatibility, position in discourse, realis match, and argument overlap.}

\subsection{Task Evaluation}
Tasks 1 and 2 are both conceptualized as binary classification and therefore a number of evaluation metrics are available. Here, I report classification accuracy\footnote{\% classified correctly}, precision\footnote{$\frac{T_p}{T_p + F_p}$}, recall\footnote{$ \frac{T_p}{T_p + F_n}$}, F$_1$-score\footnote{$F_1 = \left( 2 \cdot \frac{precision \cdot recall}{precision + recall}\right)$}, and the area under the receiver operating characteristic curve (AUC) for both tasks. Due to class imbalance, I also report BLANC scores to better capture model performance among event links and non-links \cite{recasens:hovy:2011}. The equivalent statistics, referred to as macro averaged precision, recall, and F$_1$-score, are reported for MID classification.

In out-of-sample evaluation (i.e., validation and test set performance) I use no information about the headline classes (MID incident versus non-incident) or coreferences. In other words, link predictions are conditioned on the texts and publication dates of headlines only and not on the MID status of a given headline.


\begin{table}
\begin{center}
\begin{tabularx}{\linewidth}{Xrrrrr} 
\hline\hline
\multicolumn{6}{c}{\emph{a.} MIDI classification (positive class)} \\
 & Pre. & Rec. & F$_1$ & Acc. & AUC \\ 
\cline{2-6}
Training set & 0.73 & 0.47 & 0.57 & 0.97 & 0.73 \\
Validation set & 0.89 & 0.23 & 0.36 & 0.85 & 0.61 \\
Testing set & 0.27 & 0.19 & 0.22 & 0.99 & 0.59 \\
\hline
\multicolumn{6}{c}{\emph{b.} MIDI classification (macro average)} \\
 & Pre. & Rec. & F$_1$ &  & \\ 
\cline{2-4}
Training set & 0.85 & 0.73 & 0.78 & & \\
Validation set & 0.87 & 0.61 & 0.64 & &  \\
Testing set & 0.63 & 0.59 & 0.61 & & \\
\hline
\multicolumn{6}{c}{\emph{c.} Coref prediction (positive class)} \\
 & Pre. & Rec. & F$_1$ & Acc. & AUC \\ 
\cline{2-6}
Training set & 1.00 & 0.90 & 0.95 & 0.98 & 1.00 \\
Validation set & 1.00 & 0.76 & 0.86 & 0.96 & 1.00 \\
Testing set & 0.99 & 0.45 & 0.62 & 0.91 & 1.00 \\
\hline
\multicolumn{6}{c}{\emph{d.} Coref prediction (BLANC)} \\
 & Pre. & Rec. & F$_1$ & &  \\ 
\cline{2-4}
Training set & 0.99 & 0.95 & 0.97 & &\\
Validation set & 0.98 & 0.88 & 0.92 & & \\
Testing set & 0.95 & 0.72 & 0.78 & & \\
\hline\hline
\end{tabularx}
\end{center}
\caption{Performance statistics across partitions of HoW. Positive class (\emph{a} and \emph{c}) denotes an occurrence of a MID incident or a MID incident coreference, respectively. Macro average and BLANC (\emph{b} and \emph{d})  indicate that the reported statistics have been averaged across classes with each class having been assigned equal weight.}\label{tab:results}
\end{table}

\begin{table}
\begin{center}
\begin{tabularx}{\linewidth}{rXl}
\hline \hline
MID & Pr & Headline \\ \hline
False & 0.91 & Serbs Advance in Kosovo, Imperiling... \\ 
True & 0.90 & Feuding factions meet in Congo... \\ 
True & 0.87 & Significant Rwandan troop movement ... \\ 
False & 0.87 & Serbs Stone Albanians in Divided Ko... \\ 
True & 0.86 & Zimbabwean troops deployed in Congo... \\ 
False & 0.85 & Attack in Baghdad... \\ 
False & 0.83 & Clashes in Zimbabwe... \\ 
True & 0.82 & Zimbabwe wins major battle in Congo... \\ 
True & 0.80 & Kabila moving against rebellious tr... \\ 
False & 0.80 & U.S. Cutbacks in Yemen...\\
\hline \hline
\end{tabularx}
\caption{Top ten headlines with respect to predicted probability of describing a MID.}\label{tab:topmids}
\end{center}
\end{table}

\section{Results}
I turn now to an assessment of the model's performance on both tasks: MID classification at the headline level and coreference prediction between pairs of headlines. In this analysis, only $headline_a$ results are included when assessing MID classification. This is to prevent unintentional repeat counting of headlines that appear as both $headline_a$ and $headline_b$ in different training example pairs. 

The model achieves high precision for coreference prediction but lower precision for MID classification: 0.99 and 0.27 on the testing set, respectively. Relatively high false negative rates mean that recall is low for both tasks: 0.19 for MIDI classification and 0.45 for coreference prediction.  However, considering the class imbalance present for both tasks and apparent in Table~\ref{tab:datastats}, the macro averaged or BLANC adjusted statistics are also reported. This is recommended in previous work on coreference resolution \cite{krause:etal:2016}. The model fares better for both tasks when taking this imbalance into account and achieves recall values of 0.59 and 0.72 for classification and coreference prediction, respectively. Table~\ref{tab:results} provides a full set of results for all three partitions. The final column of Table~\ref{tab:results} reports the area under the receiver operating characteristic curve (AUC). AUC can be interpreted as the probability that a randomly selected positive example will be assigned higher predicted probability of belonging to the positive class than will a randomly selected negative example. The very high accuracy and AUC scores (near 1.0) can be attributed to the high recall of the classifiers with respect to the majority negative class. The table reveals overfitting to the training set on which the model consistently achieves its highest scores.

Because content relevant to militarized interstate disputes often appears in the NYT World section, the HoW data set currently contains a significant number of false negative headlines. Table~\ref{tab:topmids} reproduces the top 10 highest scoring headlines with respect to their predicted probabilities of describing a MID. Some of the reported non-MID headlines clearly refer to MIDs.\footnote{Because these non-MID headlines are from NYT, they are not associated with a MID in HoW. I hope to reduce false negatives in future iterations of HoW.}

\begin{figure}
\begin{center}
\includegraphics[width=\linewidth]{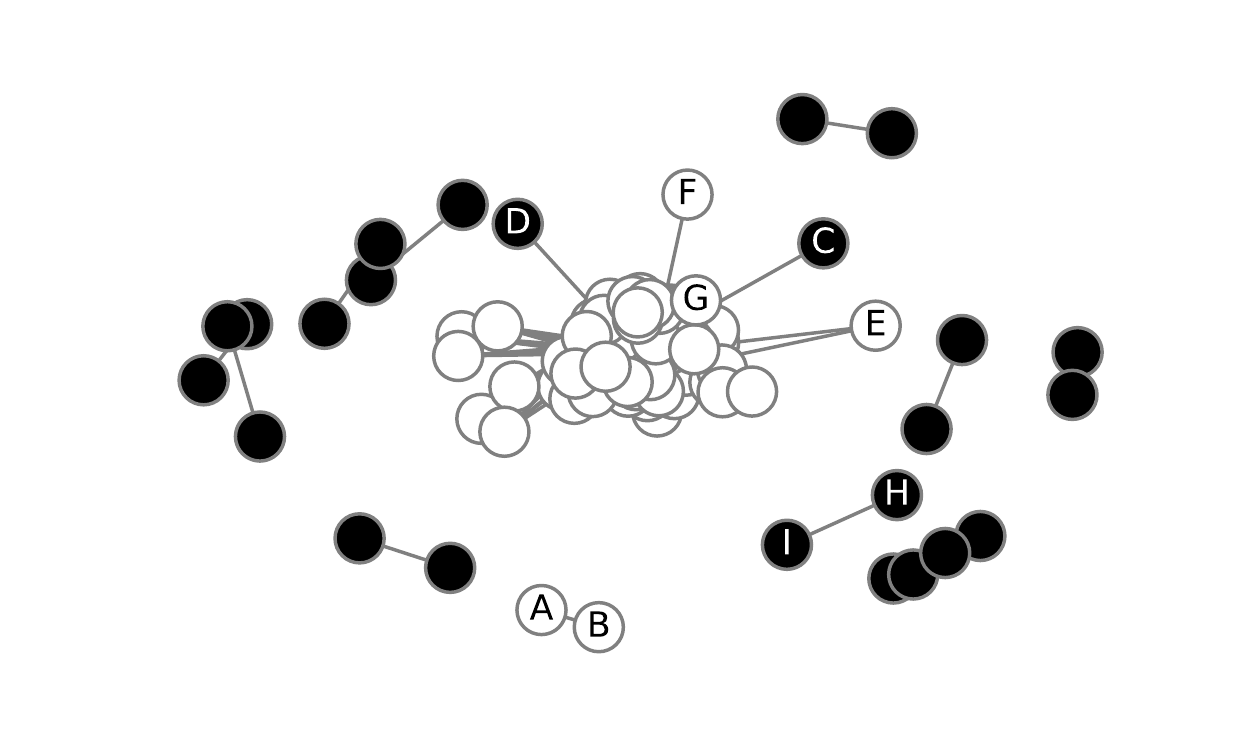}
\caption{Predicted coreferences (edges) between headlines (nodes). White nodes are true MIDS headlines; black nodes are NYT World headlines. The central cluster primarily corresponds to MID incident 4339. The \emph{A-B} pair corresponds to MID incident 4248003.}\label{fig:network}
\end{center}
\end{figure}

Figure~\ref{fig:network} depicts predicted coreferences in the test set. Two of four MID incidents are present. A selection of headlines labeled in Figure~\ref{fig:network} is provided in Table~\ref{tab:selectedheadlines}. The four MID incidents present in the HoW test set are 4248001, 4248003, 4283012, 4339, of which coreferences are identified among two or more headlines referring to 4339 and 4248003. 4339 is the Congo War. 4248001 and 4248003 are incidents between Uganda and Sudan during 1998. 4283012 is an incident between the UK and Afghanistan during the 2001 invasion of Afghanistan.

\begin{table}
\begin{center}
\begin{tabularx}{\linewidth}{Xl}
\hline\hline
ID & Headline \\
\hline
A & Sudanese plane bombed Ugandan town  aid ... \\
B & uganda condemns sudanese air attack... \\
C & One Dies as Navy Jets Collide Off Turkey... \\
D & U.S. to Change Strategy in Narcotics Fig... \\
E & Heading for an African War... \\
F & DRC gun running a rumour... \\
G & Rwanda needs and will get  a buffer zone... \\
H & Farmers Protest Against Fox in Mexico Ci... \\
I & South Koreans Challenge Northerner on U.... \\
\hline\hline
\end{tabularx}
\caption{Selected headlines from Figure~\ref{fig:network}}\label{tab:selectedheadlines}
\end{center}
\end{table}

\section{Discussion}
The HoW data set comes with a number of caveats discussed below. The negative sampling is performed by first subsetting MIDS 3.0 into the training, testing, and validation sets. Then, negative samples are picked at a rate of $5\times$ for every positive MID story pair (i.e., edge). This scale factor is selected arbitrarily and results in a  sparse graph.\footnote{While a negative sampling ratio of 5 to 1 is chosen arbitrarily, it does follow the standard in the literature for negative sampling skipgram models like word2vec \cite{mikolov:etal:2013a}.} Many negative samples describe MIDs themselves and should not be labeled as negative. No negative samples have been manually corrected and at least some false negatives can be expected. Negative samples are drawn only from the NYT World section while the MIDS 3.0 headlines are drawn from many diverse (English language) sources. Unfortunately, a representative corpus of headlines for negative sampling was unavailable at the time of writing.

Not all sources in MIDS are documented with enough specificity to identify the relevant headline. Some MID incidents only reference a section or page number and not a headline. A future step in the development of HoW will seek to identify the original source data for MID incidents that currently lack headline text to improve the coverage of MIDs over the period in question. Longer-term, additional data sources may provide event types beyond MIDs and therefore allow researchers to evaluate the out-of-class generalizability of cross-document event detection methods. In the near term, the more comprehensive headline data set for MIDS 4 (2002--2010) is being used to extend HoW and address the high proportion of missing MID incidents in HoW. 

The decision made here was to partition HoW by date. This has the advantage of offering a simple explanation of how the partitions differ from one another: they cover distinct date ranges. It also allows researchers to consider the impact of the temporal proximity of two headlines on their likelihood of being associated with the same event. In that way, date-based partitioning imitates the likely real-world scenario of cross-document event detection: near real-time monitoring. However, it also means that models fit to the training data set may generalize poorly to the testing data set since the testing data set represents events from up to five years later in time. Partitioning by time in such a way makes it difficult to control the number of positive-class observations per set. Down-sampling headlines from MIDS may help to manage partition balance but at the cost of even fewer positive MID headline examples. 

\section{Conclusion}
HoW offers a novel evaluation data set for researchers interested in automated event data and coreference resolution. Conceptualizing event data generation as a two-task problem of detection and coreference resolution will allow future efforts to better identify complex social phenomena that may otherwise be invisible given existing sentence and document-level event coding strategies. It also has implications for deduplication: the ability to automatically detect event coreferences across documents may help to reduce the number of duplicate event records that result from coverage across multiple sources.

Future efforts should seek to build on HoW by including multiple classes of events or incidents.\footnote{The previously mentioned event coreference resolution data sets contain multiple event types.} Additionally, strategies for identifying true negative samples rather than relying on the assumption that all non-MIDS headlines are negative samples will help to more precisely evaluate model performance.

\section{Bibliographical References}\label{reference}

\bibliographystyle{lrec}
\bibliography{aespen2020}

\end{document}